\documentclass[10pt,twocolumn,letterpaper]{article}
\usepackage[pagenumbers]{cvpr}
\usepackage{placeins}
\usepackage{graphicx}
\usepackage{amsmath,amssymb,booktabs,array,tabularx,multirow}
\usepackage{cuted,caption,microtype,xspace}
\usepackage{colortbl}
\usepackage{algorithm,algpseudocode}
\algrenewcommand{\algorithmicrequire}{\textbf{Input:}}
\algrenewcommand{\algorithmicensure}{\textbf{Output:}}
\definecolor{tablerowgray}{RGB}{234,232,238}
\definecolor{memblue}{RGB}{40,82,125}
\definecolor{memgray}{RGB}{244,246,248}
\newcommand{\method}{MemicVLA\xspace}
\newcommand{\piBase}{\ensuremath{\pi_{0.5}}\xspace}
\newcommand{\norm}[1]{\operatorname{norm}(#1)}

\def\arxivpublic{1}
\renewcommand{\method}{MXI-MemICL\xspace}

\definecolor{cvprblue}{rgb}{0.21,0.49,0.74}
\usepackage[pagebackref,breaklinks,colorlinks,allcolors=cvprblue]{hyperref}

\title{MXI-MemICL: Towards In-Context Robot Manipulation through Self-Retrieved Interaction Memory}
\author{%
\parbox{0.98\textwidth}{\centering\fontsize{11}{14}\selectfont
Jialong Peng\textsuperscript{1,2}, Shihao Zhou\textsuperscript{1}, Feng Yan\textsuperscript{1}, Wen Xiao\textsuperscript{1}, Haozheng Wang\textsuperscript{1}, Wenqi Ren\textsuperscript{2},\\
Wenyuan Sheng\textsuperscript{1}, Jian Qiao\textsuperscript{1}, Jiaxu Miao\textsuperscript{2,*}, Lin Ma\textsuperscript{1,*}, Chao Yang\textsuperscript{1}\\[4pt]
\fontsize{10}{12}\selectfont
\textsuperscript{1}1011.ai\quad
\textsuperscript{2}Sun Yat-sen University\\[2pt]
\textsuperscript{*}Corresponding authors\\[4pt]
\textbf{Project page:} \href{https://jialong-peng.github.io/mxi-memicl/}{MXI-MemICL}}}

\AddToHook{cmd/maketitle/after}{%
\begingroup
\setlength{\stripsep}{0pt}
\begin{strip}
\centering
\includegraphics[width=\linewidth]{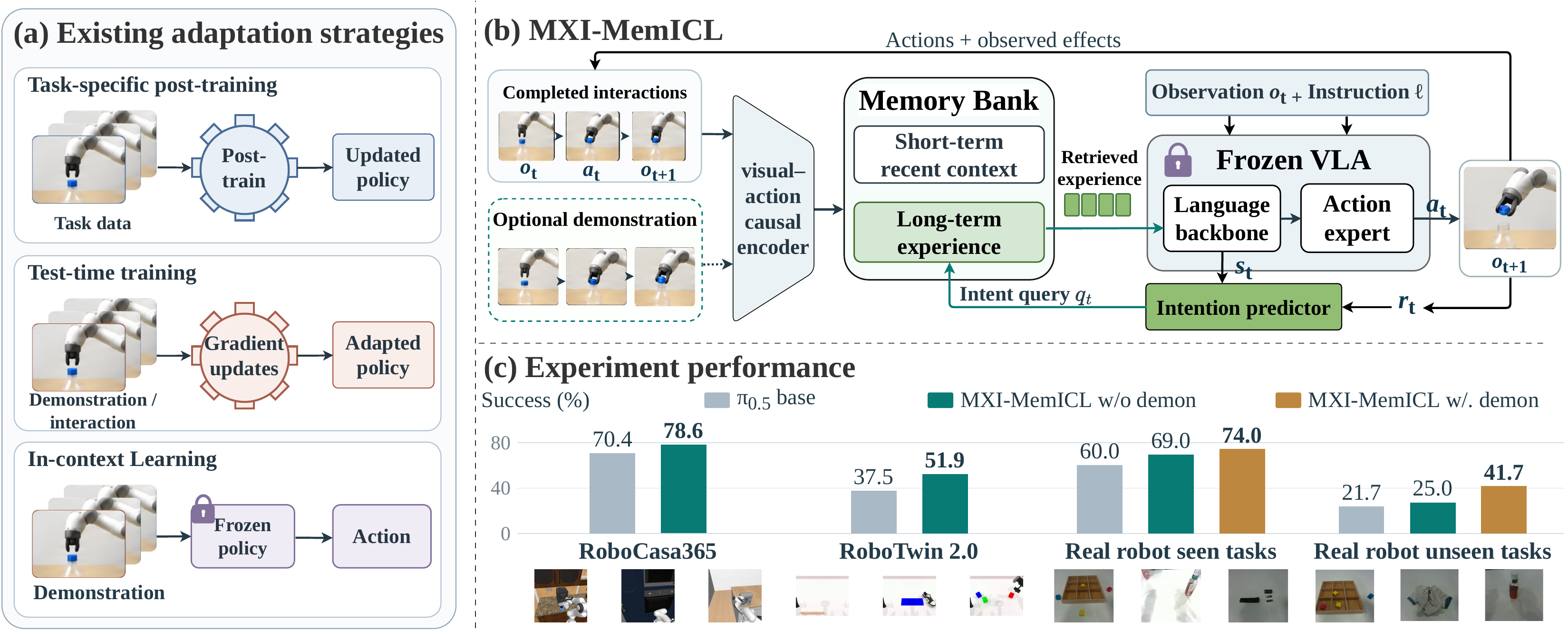}
\captionof{figure}{\looseness=-1\textbf{Experience-guided adaptation with \method.} (a) Post-training and test-time training update parameters, while ICL uses demonstrations. (b) \method uses predicted future intent to retrieve interaction or demonstration experience for a frozen VLA. (c) Success improves on RoboCasa365, RoboTwin 2.0, and real-robot tasks, with demonstrations further improving seen and unseen performance.}
\label{fig:teaser}
\par\vspace{10pt}
\end{strip}
\endgroup
}

\hypersetup{
 pdftitle={MXI-MemICL: Towards In-Context Robot Manipulation through Self-Retrieved Interaction Memory},
 pdfauthor={Jialong Peng, Shihao Zhou, Feng Yan, Wen Xiao, Haozheng Wang, Wenqi Ren, Wenyuan Sheng, Jian Qiao, Jiaxu Miao, Lin Ma, Chao Yang}
}
\begin{document}
\maketitle

\begin{abstract}
In-context learning (ICL) offers robots a way to acquire new skills from a single demonstration without parameter updates.
However, existing robotic ICL methods either rely on large-scale pretraining constrained by scarce, costly robot interaction data or require additional annotations yet struggle to generalize across tasks.
To bridge this gap, we propose \method, a framework for memory-guided ICL in vision--language--action (VLA) models.
By treating demonstrations as retrievable memory, \method enables in-context manipulation without additional large-scale pretraining or specialized annotations.
Specifically, a visual--action causal encoder turns actions and their observed effects in demonstrations and completed interactions into reusable memory to guide execution.
To retrieve experience suited to upcoming execution, an intention predictor forecasts the next stage's intent and uses it to query memory.
The retrieved experience conditions the language backbone through experience-token prefixes and gated cross-attention, guiding action generation with a learned gate that regulates its influence.
\method improves success over the baseline by +8.2\% and +14.4\% on RoboCasa365 and RoboTwin 2.0.
Real-robot experiments show a +9.0\% gain, with demonstrations adding +5.0\% on seen and +16.7\% on unseen tasks.
\label{abs:end}
\end{abstract}
\section{Introduction}
Recent advances in vision--language--action (VLA) models~\cite{rt2,pi0,pi05,openvla,grootn1,xiaomi1} and world--action models (WAM)~\cite{lingbotva,fastwam,tau0wm,cosmos3} have demonstrated impressive capabilities in diverse manipulation tasks.
However, the diversity of embodied tasks requires adapting these capabilities to new goals, objects, and environments.
Figure~\ref{fig:teaser}a summarizes three existing strategies for this adaptation.
Task-specific post-training~\cite{openvlaoft,controlvla} updates the policy using newly collected robot data, tying skill acquisition to repeated data collection and optimization.
Test-time training (TTT)~\cite{robottt,wamttt} adapts fast weights using demonstrations or interactions, but still incurs the cost of gradient optimization during deployment.

In-context learning (ICL)~\cite{icrt,bpp,instantpolicy,zerowam} offers an alternative, allowing robots to adapt from demonstrations without deployment-time parameter updates.
However, many existing approaches~\cite{skilds1,gen15} follow the GPT-3 paradigm~\cite{brown2020}, seeking emergent ICL through large-scale pretraining on diverse physical experience.
Scarce robot interaction data and high collection costs limit its scalability.
Other methods expose demonstration structure through automatically generated semantic and motion annotations~\cite{stellavla}, but generalization to more demanding unseen tasks remains challenging.

To bridge this gap, we propose \textbf{\method}, a \textbf{mem}ory-guided \textbf{i}n-\textbf{c}ontext manipulation framework for \textbf{VLA} models.
By encoding successful demonstrations as retrievable memory, \method allows a VLA to reuse demonstrated experience on new tasks.
The memory mechanism is learned from existing robot trajectories, enabling in-context manipulation without additional large-scale pretraining or specialized annotations.
At deployment, the robot adapts through experience retrieval with fixed parameters, avoiding task-specific retraining and online gradient optimization.

The language backbone's observation- and instruction-conditioned features provide semantic context for the current execution stage.
An \textbf{intention decoder} makes this context explicit as a stage embedding for experience retrieval.
To retain what was executed and its effects, we employ the causal three-stream architecture of BehaviorVLA~\cite{behaviorvla} as a \textbf{visual--action causal encoder}, encoding actions and observed visual changes into memory.
\method stores memories with context-dependent retrieval keys in a unified \textbf{memory bank}, organized by recency into short-term and long-term memory.
Short-term memory maintains ongoing execution context, while long-term memory retains earlier experience for reuse.

The relevant experience changes as execution progresses, even within a single demonstration.
Retrieval must therefore account for both the observed stage and the intent of the upcoming action.
To select experience that can guide the next stage, an \textbf{intention predictor} anticipates next-stage intent from experience-conditioned backbone features.
The predicted intent combines with the observed stage embedding to retrieve experience for upcoming actions from prior interactions and any supplied demonstration.
Recent context and retrieved experience are fused into \textbf{experience tokens} that condition the language backbone through prefixes and \textbf{gated cross-attention}.
The learned gates regulate the strength of this guidance during action generation.
Because demonstrations and online interactions share a memory representation, the robot reuses demonstrated experience through the same retrieval and injection mechanism.

We evaluate \method on real robots, RoboCasa365~\cite{robocasa365}, RoboTwin 2.0~\cite{robotwin2}, and LIBERO~\cite{libero}.
Relative to $\pi_{0.5}$, \method improves success by +8.2\% on RoboCasa365 and +14.4\% on the reported RoboTwin 2.0 task subset.
On real robots, it ranks first among the compared methods on all five tasks, with an average gain of +9.0\% over the base policy.
A single successful demonstration adds +5.0\% on seen tasks and +16.7\% on unseen tasks over the no-demonstration policy, enabling experience-guided adaptation without parameter updates at deployment.

Our contributions are summarized as follows:
\begin{itemize}
\item We introduce \method, which turns demonstrations and interaction trajectories into reusable memory for demonstration-guided VLA adaptation without parameter updates at deployment.
\item We develop intention-guided retrieval and two-stream experience injection to select and incorporate experience relevant to upcoming actions.
\item Extensive experiments in simulation and on real robots demonstrate the superiority of \method's memory architecture, with ablations and analyses further validating the contributions of individual components.
\end{itemize}

\begin{figure*}[t]
\centering
\includegraphics[width=\linewidth]{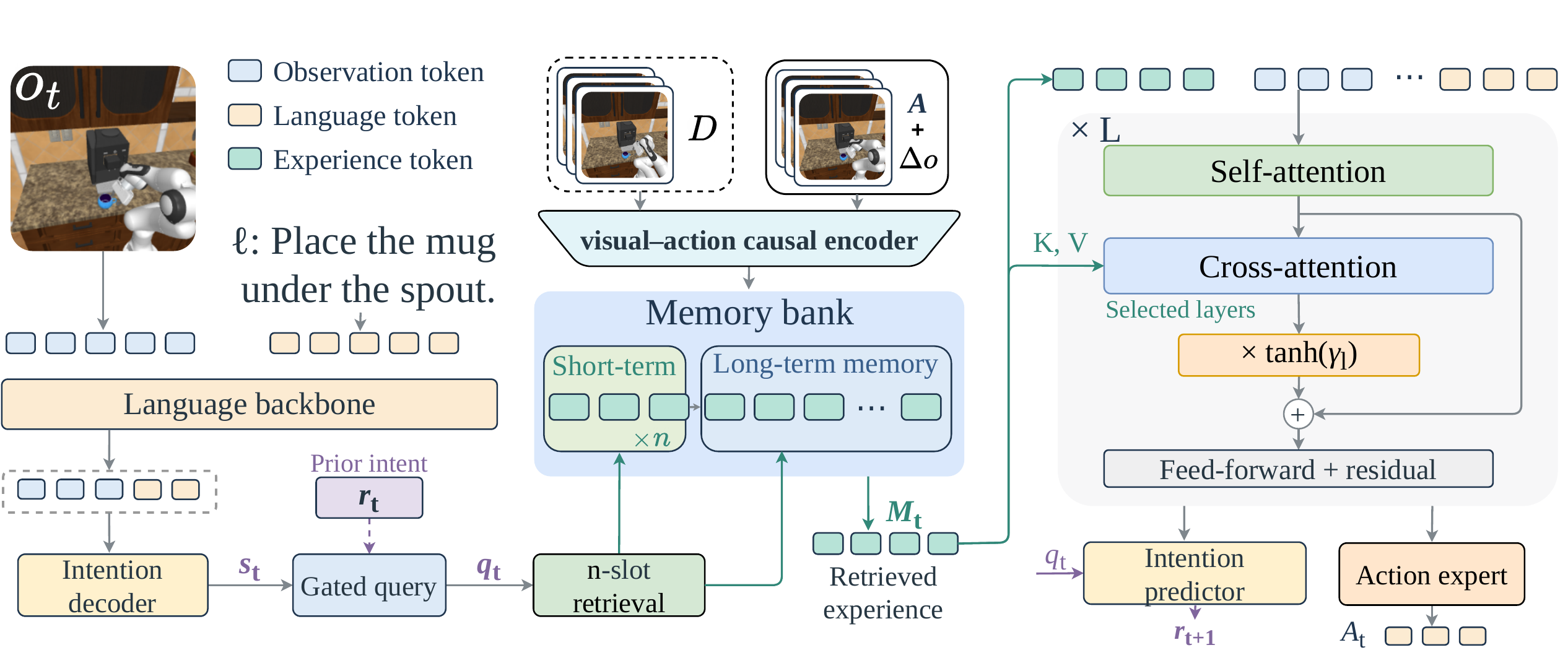}
\caption{\textbf{Overall architecture of \method.} Stage embeddings from observations and instructions combine with predicted intent for retrieval. A visual--action causal encoder encodes completed interactions and optional demonstrations into memory. Retrieved experience and recent context form $n$ experience tokens that condition the shared backbone via prefixes and gated cross-attention, whose learned gates regulate injection strength. Backbone outputs support action generation and next-step intent prediction. The intention decoder and intention predictor use residual MLPs. \textbf{Detailed diagrams of all modules} are provided in the supplementary material.}
\label{fig:method}
\end{figure*}

\section{Related Work}
\paragraph{VLA and WAM Foundation Models.}
VLA and WAM foundation models advance general-purpose manipulation by connecting multimodal understanding and dynamics modeling with action generation.
RT-2~\cite{rt2} and OpenVLA~\cite{openvla} predict discrete action tokens, while $\pi_0$~\cite{pi0} and GR00T N1~\cite{grootn1} generate continuous actions.
\piBase~\cite{pi05} and Xiaomi-Robotics-1~\cite{xiaomi1} further improve generalization through training on broader datasets.
LingBot-VA~\cite{lingbotva}, $\tau_0$-WM~\cite{tau0wm}, and Cosmos 3~\cite{cosmos3} connect world modeling with control, while Fast-WAM~\cite{fastwam} avoids test-time video generation.
However, adapting to new tasks often requires task-specific data and post-training~\cite{openvlaoft,controlvla}.
To address this limitation, \method introduces a demonstration-memory interface for reusing existing atomic skills without parameter updates.

\paragraph{In-Context Learning in Large Language Models.}
GPT-3~\cite{brown2020} demonstrates that language models can perform tasks using in-context examples without parameter updates.
Subsequent work examines what demonstrations convey, identifying the roles of label space, input distribution, and prompt format~\cite{min2022}.
Mechanistic studies interpret ICL as implicit Bayesian inference~\cite{xie2021}, examine function learning from context~\cite{garg2022}, and show that transformers can implement regression algorithms within a forward pass in controlled settings~\cite{akyurek2023,vonoswald2022}.
These studies motivate learning through context.
\method focuses on organizing relevant experience into context for action generation.

\paragraph{Robotic In-Context Learning Methods.}
Robotic ICL methods differ in how they encode and use demonstrations: VIMA~\cite{vima} uses multimodal prompts, ICRT~\cite{icrt} models sensorimotor sequences, and Instant Policy~\cite{instantpolicy} uses graph diffusion.
Behavior Prompting Policy~\cite{bpp} emphasizes task diversity and sensorimotor detail, while Zero-WAM~\cite{zerowam} and StellaVLA~\cite{stellavla} use human videos and automatically annotated demonstrations, respectively.
Zeva~\cite{zeva} encodes actions and their effects into retrievable memory for a frozen policy.
\method retrieves experience using predicted next-stage intent and uses the same query to retrieve demonstration memory, offering a new avenue toward robotic ICL.

\section{Method}
\label{sec:method}
\label{sec:memory}
In this section, we first formulate the problem in Section~\ref{sec:formulation}, then describe stage and memory generation in Section~\ref{sec:generation}, experience retrieval in Section~\ref{sec:retrieval}, and experience injection in Section~\ref{sec:injection}.
Section~\ref{sec:training} presents our training strategy.
Figure~\ref{fig:method} provides an overview of our method.

\subsection{Problem Formulation}
\label{sec:formulation}
Given a language instruction $\ell$ and observation $o_t=(I_t^1,\ldots,I_t^N,v_t)$, a VLA policy $\pi_\theta(A_t\mid o_t,\ell)$ predicts $H$ successive controls $A_t=(a_{t,0},\ldots,a_{t,H-1})\in\mathbb R^{H\times d_a}$ at decision $t$~\cite{pi05}.
The observation contains one or more camera images and proprioceptive state $v_t$, and each control has $d_a$ action dimensions.
\method further conditions action generation on memory from completed interactions and an optional successful demonstration $D$.
With $x_t=(o_t,\ell)$, the current stage embedding $s_t$ and prior intent $r_t$ form the gated query $q_t=G_\phi(s_t,r_t)$.
Query formation and experience retrieval condition action generation:
\begin{equation}
\begin{aligned}
 &\pi_{\theta,\phi}(A_t\mid x_t,\mathcal B_t,r_t)
 =\iint\underbrace{\pi_{\theta,\phi}(A_t\mid x_t,M_t)}_{\text{Action generation}}\\
 &\quad\cdot\underbrace{p_\phi^{\rm ret}(M_t\mid q_t,\mathcal B_t)}_{\text{Experience retrieval}}
 \underbrace{p_\phi^{\rm qry}(q_t\mid s_t,r_t)}_{\text{Query formation}}
 \,\mathrm dM_t\,\mathrm dq_t,
\end{aligned}
\label{eq:policy}
\end{equation}
which grounds robot behavior in relevant experience.
The \textbf{memory bank} $\mathcal B_t$ stores key--memory pairs.
Query formation $p_\phi^{\rm qry}$ combines $s_t$ and $r_t$ into $q_t$, while experience retrieval $p_\phi^{\rm ret}$ retrieves and fuses experience from $\mathcal B_t$ into \textbf{experience tokens} $M_t$.

\subsection{Stage and Memory Generation}
\label{sec:generation}
\method encodes the current stage to form queries and completed interactions to build indexed memories (Figure~\ref{fig:method}).
The backbone feature extractor $F_\theta$ produces hidden representations $\mathbf H_t=F_\theta(o_t,\ell)$.
Mean pooling yields $h_t=\operatorname{Pool}(\mathbf H_t)$, which the \textbf{intention decoder} maps to a normalized stage embedding $s_t=D_\phi(h_t)$.

In parallel, the interaction memory $m_t$ comes from a \textbf{visual--action causal encoder} based on the causal three-stream architecture~\cite{behaviorvla,zeva}.
Its visual, action, and interaction-state streams process executed controls together with observations before and after execution, using recurrent updates and attention across streams.
Fusing the interaction states, visual changes, and an action summary produces a memory of what was executed during the stage and its observed effects.
We index this memory with a retrieval key $k_t$, generated by an MLP from the pooled feature $h_t$ and pre-execution interaction state $z_t$.

Key-memory pairs from completed interactions enter the shared \textbf{memory bank} $\mathcal B_t$, which holds up to 64 pairs.
The current attempt's latest $n$ records provide short-term execution context $\mathcal B_t^{\rm short}$, while the remaining records form long-term memory $\mathcal B_t^{\rm long}$.
These disjoint views update as new interactions enter the bank.

\begin{algorithm}[!t]
\caption{\method Inference}
\label{alg:inference}
\footnotesize
\begin{algorithmic}[1]
\Require Instruction $\ell$, environment $\mathcal E$, horizon $T$, execution length $h$, context count $n$
\Require Optional successful demonstration $D$ ($\varnothing$ if absent)
\Ensure Executed actions and updated online memory $\mathcal B^E$
\State Freeze all model parameters $\theta,\phi,\eta$
\State $\mathcal B^E\gets\varnothing$, $\mathcal B^D\gets\varnothing$, $r_0\gets\bot$
\If{$D\ne\varnothing$} \Comment{Independent demo encoder state}
  \State $\mathcal B^D\gets\operatorname{EncodeDemo}(D,\ell)$
\EndIf
\State $o_0\gets\operatorname{Reset}(\mathcal E)$, $z_0\gets\operatorname{InitC}(o_0)$
\For{$t=0,\ldots,T-1$}
  \State $\mathbf H_t\gets F_\theta(o_t,\ell)$ \Comment{Backbone pass 1}
  \State $h_t\gets\operatorname{Pool}(\mathbf H_t)$, $s_t\gets D_\phi(h_t)$
  \State $q_t\gets G_\phi(s_t,r_t)$ \Comment{$q_0=s_0$}
  \State $(\mathcal B_t^{\rm short},\mathcal B_t^{\rm long})\gets\operatorname{SplitByRecency}_n(\mathcal B^E)$
  \State $\mathcal C_t\gets\mathcal B^D\cup\mathcal B_t^{\rm long}$ \Comment{Full demo, causal execution}
  \State $\widetilde{\mathcal M}_t\gets\operatorname{Retrieve}_n(q_t,\mathcal C_t)$ \Comment{$n$ soft slots}
  \State $M_t\gets\operatorname{FuseExperience}_n(q_t,\mathcal B_t^{\rm short},\widetilde{\mathcal M}_t)$
  \State $(\bar h_t,\mathrm{KV}_t)\gets\operatorname{BackboneWithExperience}(o_t,\ell,M_t)$
  \State $r_{t+1}\gets\operatorname{PredictIntent}(\bar h_t,q_t)$ \Comment{Next decision}
  \State $A_t\gets\operatorname{ActionExpert}(\mathrm{KV}_t)$ \Comment{Flow matching}
  \State $(A_t^{\rm exec},o_{t+1},d_t)\gets\operatorname{Execute}(\mathcal E,A_t[1\!:\!h])$
  \State $(z_{t+1},e_t)\gets C_\eta(z_t,o_t,A_t^{\rm exec},o_{t+1})$
  \State $k_t\gets\operatorname{MLP}_{\rm key}(h_t,z_t)$ \Comment{Normalized key}
  \State $m_t\gets\operatorname{EncodeMemory}([z_t,z_{t+1},e_t,\operatorname{Summary}(A_t^{\rm exec})])$
  \State $\mathcal B^E\gets\operatorname{CommitCompleted}(\mathcal B^E,(k_t,m_t))$
  \State \textbf{if} $d_t$ \textbf{then break}
\EndFor
\State \Return $\mathcal B^E$
\end{algorithmic}
\end{algorithm}

\subsection{Intention-Guided Experience Retrieval}
\label{sec:retrieval}
At each decision, the stage embedding $s_t$ summarizes the current observation and instruction, while $r_t$ carries the intent predicted for this step at the preceding decision.
The gated query mapping $G_\phi$ combines these cues through a learned gate $\alpha_t$ and $\ell_2$ normalization, with $q_0=s_0$:
\begin{equation}
 q_t=G_\phi(s_t,r_t)=\norm{s_t+\alpha_t(r_t-s_t)}.
 \label{eq:query}
\end{equation}

\paragraph{Experience retrieval.}
The query $q_t$ selects experience from long-term memory $\mathcal B_t^{\rm long}$ through $n$ soft retrieval slots.
Each slot adds a learned query offset, computes cosine similarities between projected queries and keys, and aggregates memory values with normalized weights $w_{r,i}$:
\begin{equation}
 \tilde m_{t,r}=\sum_{i\in\mathcal B_t^{\rm long}}w_{r,i}m_i,
 \qquad r=1,\ldots,n.
 \label{eq:retrieval}
\end{equation}
Records assigned high weights become less available to later slots, promoting complementary retrieval.
The retrieved experience $\widetilde{\mathcal M}_t$ is fused with recent execution context into experience tokens for backbone injection.

\paragraph{Memory-based in-context adaptation.}
\label{sec:stage3}
To support robotic ICL, \method turns a single successful demonstration into retrievable experience that guides action generation without parameter updates at deployment.
The demonstration $D$ and target execution $E$ come from distinct source trajectories of the same task and are encoded separately by the causal encoder.
All completed demonstration records form the read-only bank $\mathcal B^D$, while execution memory $\mathcal B^E_{<t}$ contains interactions completed before decision $t$.
The latest $n$ execution records form the short-term context, while the long-term candidates combine the full demonstration with the remaining execution records:
\begin{equation}
 \mathcal C_t=\mathcal B^D\cup
 (\mathcal B^E_{<t}\setminus\mathcal B_t^{\rm short}).
 \label{eq:demo-memory}
\end{equation}
Execution initializes its own intent and encoder states from its initial observation and instruction.
The recurrent query then retrieves experience relevant to the anticipated stage from $\mathcal C_t$ through the same $n$-slot soft retriever.
The retrieved demonstration experience is fused with recent execution context and injected into the language backbone to guide subsequent actions.
Inference with and without a demonstration follows Algorithm~\ref{alg:inference}.

\setlength{\textfloatsep}{12pt plus 2pt minus 2pt}
\setlength{\dbltextfloatsep}{12pt plus 2pt minus 2pt}
\begin{table*}[t]
\centering
\captionsetup{singlelinecheck=false}
\caption{\textbf{Performance comparison on RoboTwin 2.0.} Per-task success rates (\%) over 100 rollouts under the Hard setting. $\dagger$ denotes our reproduced result. Other baselines are from the leaderboard. Results on additional 12 tasks are provided in the supplementary material.}
\label{tab:robotwin}
{\scriptsize
\setlength{\tabcolsep}{1.5pt}
\begin{tabularx}{\linewidth}{@{}l|>{\centering\arraybackslash}X>{\centering\arraybackslash}X>{\centering\arraybackslash}X>{\centering\arraybackslash}X>{\centering\arraybackslash}X>{\centering\arraybackslash}X>{\centering\arraybackslash}X>{\centering\arraybackslash}X>{\centering\arraybackslash}X>{\centering\arraybackslash}X@{}}
\toprule
\multicolumn{1}{@{}l|}{\multirow{2}{*}{\textbf{Method}}} & \multicolumn{10}{c}{\textbf{RoboTwin 2.0 (Hard)}} \\
\cmidrule(lr){2-11}
 & \shortstack{Grab\\Roller} & \shortstack{Hammer\\Block} & \shortstack{Move Card\\Away} & \shortstack{Place A2B\\Right} & \shortstack{Place\\Shoe} & \shortstack{Put Object\\in Cabinet} & \shortstack{Rank Blocks\\(RGB)} & \shortstack{Scan\\Object} & \shortstack{Stack\\2 Blocks} & \shortstack{Stack\\3 Bowls} \\
\midrule
Fast-WAM~\cite{fastwam} & 2 & 0 & 0 & 1 & 0 & 0 & 0 & 0 & 0 & 0 \\
Discrete Forcing~\cite{discreteforcing} & 36 & 4 & 4 & 2 & 6 & 0 & 0 & 4 & 8 & 4 \\
$\pi_0$~\cite{pi0} & 80 & 21 & 22 & 6 & 6 & 18 & 5 & 1 & 1 & 24 \\
X-VLA~\cite{xvla} & 20 & 9 & 5 & 13 & 39 & 21 & 4 & 9 & 2 & 9 \\
Xiaomi-Robotics-0~\cite{xiaomi0} & 12 & 16 & 12 & 2 & 29 & 5 & 8 & 8 & 22 & 35 \\
EventVLA~\cite{eventvla} & 48 & 1 & 17 & 17 & 10 & 27 & 6 & 3 & 13 & 11 \\
$\pi_{0.5}$$^{\dagger}$ ~\cite{pi05} & 83 & 14 & 20 & 29 & 57 & 23 & 39 & 37 & 27 & 46 \\
\midrule
\rowcolor{tablerowgray}
\multicolumn{1}{@{}l|}{\textbf{\method{} (Ours)}} & \textbf{96} & \textbf{22} & \textbf{28} & \textbf{40} & \textbf{72} & \textbf{38} & \textbf{57} & \textbf{51} & \textbf{62} & \textbf{53} \\
\bottomrule
\end{tabularx}

\par}
\end{table*}

\begin{table}[t]
\centering
\captionsetup{singlelinecheck=false}
\caption{\textbf{Performance comparison on RoboCasa365.} Per-task SR (\%) over 50 rollouts per run. All results are reproduced by us. Eight additional tasks are reported in the supplementary material.}
\label{tab:atomic}
{\fontsize{6.5}{7.5}\selectfont
\setlength{\tabcolsep}{0.7pt}
\begin{tabularx}{\linewidth}{@{}p{68pt}|>{\hsize=1.08\hsize\linewidth=\hsize\centering\arraybackslash}X>{\hsize=1.0\hsize\linewidth=\hsize\centering\arraybackslash}X>{\hsize=0.9\hsize\linewidth=\hsize\centering\arraybackslash}X>{\hsize=1.0\hsize\linewidth=\hsize\centering\arraybackslash}X>{\hsize=1.02\hsize\linewidth=\hsize\centering\arraybackslash}X|r@{}}
\toprule
\textbf{Method} & {\fontsize{5.2}{6.5}\selectfont\bfseries\shortstack{TurnOn\\ElectricKettle}} & {\fontsize{5.2}{6.5}\selectfont\bfseries\shortstack{CloseToaster\\OvenDoor}} & {\fontsize{5.2}{6.5}\selectfont\bfseries\shortstack{TurnOn\\Microwave}} & {\fontsize{5.2}{6.5}\selectfont\bfseries\shortstack{CoffeeSetup\\Mug}} & {\fontsize{5.2}{6.5}\selectfont\bfseries\shortstack{OpenStand\\MixerHead}} & \textbf{Avg.} \\
\midrule
\shortstack[l]{Xiaomi-Robotics-1~\cite{xiaomi1}} & 80.0 & 76.0 & 62.0 & 44.0 & 92.0 & 70.8 \\
Fast-WAM~\cite{fastwam} & 76.0 & 72.0 & 36.0 & 20.0 & 78.0 & 56.4 \\
Cosmos3-Nano~\cite{cosmos3} & 90.0 & 68.0 & 26.0 & 16.0 & 80.0 & 56.0 \\
LingBot-VA~\cite{lingbotva} & 82.0 & 48.0 & 10.0 & 6.0 & 86.0 & 46.4 \\
$\tau_0$-WM~\cite{tau0wm} & 68.0 & 30.0 & 20.0 & 38.0 & 70.0 & 45.2 \\
$\pi_{0.5}$~\cite{pi05} & 84.0 & 80.0 & \textbf{64.0} & 30.0 & 94.0 & 70.4 \\
Zeva~\cite{zeva} & 93.0 & \textbf{88.0} & 57.0 & 44.0 & 94.0 & 75.2 \\
\midrule
\rowcolor{tablerowgray}
\multicolumn{1}{@{}p{68pt}|}{\shortstack[l]{\method{}\ifdefined\arxivpublic\\(Ours)\else\ (Ours)\fi}} & \textbf{94.0} & 86.0 & \textbf{64.0} & \textbf{53.0} & \textbf{96.0} & \multicolumn{1}{|r@{}}{\textbf{78.6}} \\
\bottomrule
\end{tabularx}

\par}
\end{table}

\begin{table*}[t]
\centering
\captionsetup{singlelinecheck=false}
\caption{\textbf{Performance comparison on real-robot tasks.} We report per-task SR over 20 rollouts on ARX~X5. All results are reproduced by us. \method outperforms the \piBase baseline and other models on both short- and long-horizon tasks.}
\label{tab:real}
{\small
\setlength{\tabcolsep}{2.5pt}
\begin{tabularx}{\linewidth}{@{}l|>{\centering\arraybackslash}X>{\centering\arraybackslash}X|>{\centering\arraybackslash}X>{\centering\arraybackslash}X>{\centering\arraybackslash}X|>{\centering\arraybackslash}X@{}}
\toprule
\multirow{2}{*}[-6pt]{Method} & \multicolumn{2}{c|}{\textbf{Short-horizon tasks}} & \multicolumn{3}{c|}{\textbf{Long-horizon tasks}} & \multirow{2}{*}[-6pt]{Average} \\
\cmidrule(lr){2-3}\cmidrule(lr){4-6}
& \shortstack{Block\\induction} & \shortstack{Water\\pouring} & \shortstack{Clothes\\folding} & \shortstack{Bottle-cap\\twisting} & \shortstack{Battery\\insertion} & \\
\midrule
Cosmos3-Nano~\cite{cosmos3} & 40\% & 25\% & 20\% & 0\% & 0\% & 17.0\% \\
Fast-WAM~\cite{fastwam} & 70\% & 75\% & 20\% & 15\% & 10\% & 38.0\% \\
$\tau_0$-WM~\cite{tau0wm} & 85\% & 65\% & 25\% & 5\% & 5\% & 37.0\% \\
Xiaomi-Robotics-1~\cite{xiaomi1} & 60\% & 85\% & 40\% & 25\% & 5\% & 43.0\% \\
$\pi_{0.5}$~\cite{pi05} & 85\% & 85\% & 60\% & 40\% & 30\% & 60.0\% \\
\midrule
\rowcolor{tablerowgray}
\multicolumn{1}{@{}l|}{\method{} (Ours)} & \textbf{95\%} & \textbf{90\%} & \textbf{70\%} & \textbf{45\%} & \textbf{45\%} & \multicolumn{1}{|c@{}}{\textbf{69.0\%}} \\
\bottomrule
\end{tabularx}

\par}
\bigskip
\centering
\includegraphics[width=\linewidth]{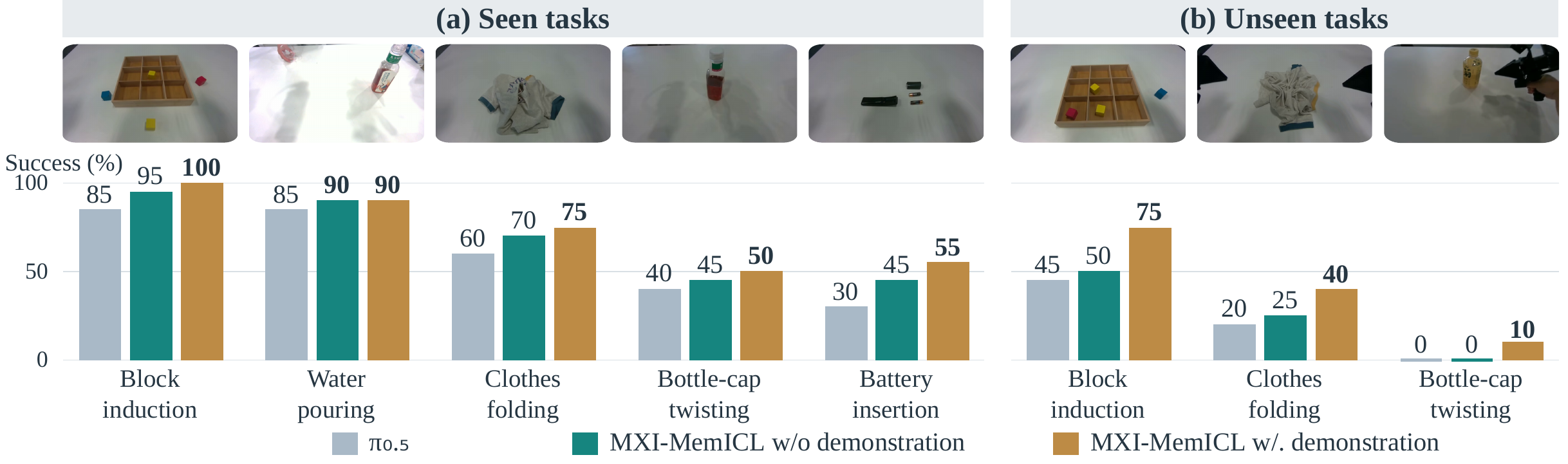}
\captionof{figure}{\textbf{Real-robot in-context learning on seen and unseen tasks.} Panels (a) and (b) show seen and unseen tasks, respectively. Bars compare $\pi_{0.5}$, \method without a demonstration, and \method with one successful demonstration. Numbers denote success rates (\%).}
\label{fig:real-icl}
\end{table*}

\subsection{Two-Stream Experience Injection}
\label{sec:injection}
We fuse recent execution context and retrieved experience into $n$ experience tokens $M_t$.
$n$ learned queries, conditioned on $q_t$, attend separately to recent experience from short-term memory and retrieved experience through two attention branches.
The two attention outputs are concatenated, fused, and projected to the backbone's feature width.
The same fusion incorporates retrieved demonstration experience when available.

We prepend these experience tokens to the observation and instruction tokens and further incorporate them through \textbf{gated cross-attention}.
Cross-attention updates the hidden representations $\mathbf H_{t,l}$ at selected layers $l$ with a residual derived from the experience tokens:
\begin{equation}
 \mathbf H_{t,l}'=\mathbf H_{t,l}+\tanh(\gamma_l)
 \operatorname{Attn}_l(\operatorname{RMSNorm}(\mathbf H_{t,l}),M_t).
 \label{eq:injection}
\end{equation}
The learnable gates $\gamma_l$ are initialized to zero.
The learned coefficients $\tanh(\gamma_l)$ regulate how strongly the injected experience influences each layer's hidden representations.
The action expert generates action chunks from the experience-conditioned backbone key--value cache, reusing it throughout flow-matching sampling.
The same forward pass yields the conditioned readout $\bar h_t$ by mean pooling over the final hidden states of the experience-token prefixes.
The \textbf{intention predictor} combines this readout with the current query $q_t$ through a learned residual mapping $f_{\rm next}$ to produce the next intent prediction:
\begin{equation}
 r_{t+1}=\norm{q_t+f_{\rm next}(\bar h_t,q_t)}.
 \label{eq:next-intent}
\end{equation}

\subsection{Training Strategy}
\label{sec:training}
After base-policy preparation, the three training phases learn interaction representations, experience-conditioned control, and demonstration-guided retrieval.
\vspace{-2pt}

\paragraph{Phase 1: interaction encoding.}
We train the causal encoder to predict executed actions and subsequent visual features, preserving both behavior and its effects in the interaction representation.
Auxiliary objectives group same-task experience through contrastive learning, preserve temporal progression through phase regularization, and associate actions with observed changes through effect supervision.
The resulting encoder is frozen in subsequent stages to provide a shared interaction representation for memory.
\vspace{-2pt}

\paragraph{Phase 2: experience-conditioned control.}
With the base policy frozen, we jointly optimize intent modeling and experience retrieval, fusion, and injection.
Action supervision uses flow matching~\cite{flowmatching} with intent and anchor objectives:
\begin{equation}
 \mathcal L_{\rm II}=\mathcal L_{\rm FM}
 +\lambda_I\mathcal L_{\rm intent}+\lambda_A\mathcal L_{\rm anchor}.
 \label{eq:policy-loss}
\end{equation}
A fixed, normalized projection of $h_t$ supplies the semantic anchor $u_t$.
The intent loss aligns the predicted prior $r_{t+1}$ with the next anchor $u_{t+1}$:
\begin{equation}
 \mathcal L_{\rm intent}=\mathbb E[1-\cos(r_{t+1},\operatorname{sg}(u_{t+1}))].
 \label{eq:intent-loss}
\end{equation}
Complementing this prediction, the anchor loss grounds the retrieval query $q_t$ in the current observation:
\begin{equation}
 \mathcal L_{\rm anchor}=\mathbb E[1-\cos(q_t,\operatorname{sg}(u_t))].
 \label{eq:anchor-loss}
\end{equation}
Expectations are over valid supervised samples, $\operatorname{sg}$ stops target gradients, and $\lambda_I,\lambda_A$ weight the two auxiliary losses.
Training across consecutive decisions also lets downstream action loss refine predicted intent through retrieval.
\vspace{-2pt}

\paragraph{Phase 3: demonstration-guided adaptation.}
Phase 3 starts from Phase 2 weights and uses independent same-task demonstration--execution pairs with their native task instructions.
The backbone and causal encoder remain frozen, while the intent and memory parameters trained in Phase 2 continue to adapt.
Target actions are supervised with the Phase 2 objective.
Future execution observations supply intent labels only. Retrieval, recovery, and action-prior auxiliaries are disabled.
Implementation details and complete training objectives are provided in the supplementary material.

\vspace{-2pt}
\section{Experiments}
\label{sec:experiments}
We evaluate the effectiveness of our closed-loop memory architecture and memory-guided in-context manipulation through simulation benchmarks and real-robot experiments.
Further efficiency analyses and visualizations are provided in the supplementary material.

\subsection{Evaluation on Simulation Benchmarks}
\paragraph{Simulation Benchmarks.}
We evaluate \method on RoboTwin 2.0~\cite{robotwin2}, RoboCasa365~\cite{robocasa365}, and LIBERO~\cite{libero}, covering multi-stage manipulation, atomic kitchen skills, and diverse task suites.
On RoboTwin 2.0, we evaluate 22 tasks over 100 episodes each under the Hard setting with scene randomization.
For RoboCasa365, we select 13 tasks covering object placement and articulated-object manipulation and evaluate each task on 50 fixed scenes with one attempt per scene.
Detailed LIBERO settings, results, and analysis are provided in the supplementary material.
For all benchmarks, we report per-task success rates (SR) and their equally weighted mean.
We evaluate each simulation experiment at least twice.

\paragraph{Implementation Details.}
We build \method on \piBase~\cite{pi05} with memory encoding, intention-guided retrieval, and two-stream experience injection.
By default, we train the memory components for 8k steps in Phase 2 and 4k in Phase 3 on eight NVIDIA PRO 6000 GPUs with a batch size of 64.
We use $n=4$ unless otherwise specified.
Training configurations and deployment details are provided in the supplementary material.

\paragraph{Results on RoboTwin 2.0.}
On the RoboTwin 2.0 Hard tasks in Table~\ref{tab:robotwin}, \method achieves a mean success rate of 51.9\% versus 37.5\% for \piBase, outperforming all listed baselines on every task.
Specifically, success rises from 27\% to 62\% on Stack 2 Blocks and from 39\% to 57\% on Rank Blocks.
Both tasks involve successive object interactions in which each completed interaction changes the next required action.
Intention-guided retrieval supplies action--effect experience for such transitions without requiring the current observation to encode the entire execution history.
These results show that anticipating next-stage intent enables \method to retrieve relevant experience from the memory bank to guide subsequent action generation.

\paragraph{Results on RoboCasa365.}
On the five tasks in Table~\ref{tab:atomic}, \method achieves 78.6\% mean success, exceeding 70.4\% for \piBase and 70.8\% for Xiaomi-Robotics-1.
Using the same \piBase base model and training and evaluation settings, our reproduction of the robotic ICL method Zeva~\cite{zeva} achieves a success rate of 75.2\%, lower than that of \method.
Compared with the \piBase baseline, our method raises Coffee Setup Mug success from 30\% to 53\%.
\method also achieves 94\%, 86\%, and 96\% on Electric Kettle, Toaster Oven Door, and Mixer Head while maintaining 64\% on Microwave.
The largest improvement occurs where base execution is least reliable, rather than on the nearly saturated Mixer Head task.
This pattern suggests that the benefits of experience-conditioned manipulation vary across tasks, with harder tasks relying more on experience guidance and showing larger gains for \method.

\begin{table}[t]
\centering
\captionsetup{singlelinecheck=false}
\caption{\textbf{Memory-timescale ablation on RoboCasa365.} We report per-task SR (\%) over 50 rollouts. }
\label{tab:timescale}
{\fontsize{6.5}{7.5}\selectfont
\setlength{\tabcolsep}{1pt}
\begin{tabularx}{\linewidth}{@{}cc|>{\centering\arraybackslash}X>{\centering\arraybackslash}X>{\centering\arraybackslash}X>{\centering\arraybackslash}X>{\centering\arraybackslash}X|r@{}}
\toprule
\multicolumn{2}{@{}c|}{\textbf{Method}} & \multirow{2}{*}{{\fontsize{5.5}{6.5}\selectfont\bfseries\shortstack{TurnOn\\ElectricKettle}}} & \multirow{2}{*}{{\fontsize{5.5}{6.5}\selectfont\bfseries\shortstack{CloseToaster\\OvenDoor}}} & \multirow{2}{*}{{\fontsize{5.5}{6.5}\selectfont\bfseries\shortstack{TurnOn\\Microwave}}} & \multirow{2}{*}{{\fontsize{5.5}{6.5}\selectfont\bfseries\shortstack{CoffeeSetup\\Mug}}} & \multirow{2}{*}{{\fontsize{5.5}{6.5}\selectfont\bfseries\shortstack{OpenStand\\MixerHead}}} & \multirow{2}{*}{\textbf{Avg.}} \\
\cmidrule(lr){1-2}
\textbf{Short} & \textbf{Long} &  &  &  &  &  &  \\
\midrule
 &  & 84.0 & 80.0 & \textbf{64.0} & 30.0 & 94.0 & 70.4 \\
$\checkmark$ &  & 92.0 & 77.0 & 63.0 & 46.0 & 94.0 & 74.4 \\
 & $\checkmark$ & \textbf{94.0} & 78.0 & 58.0 & 38.0 & \textbf{96.0} & 72.8 \\
\midrule
\rowcolor{tablerowgray}
$\checkmark$ & \multicolumn{1}{c|}{$\checkmark$} & \textbf{94.0} & \textbf{86.0} & \textbf{64.0} & \textbf{53.0} & \textbf{96.0} & \multicolumn{1}{|r@{}}{\textbf{78.6}} \\
\bottomrule
\end{tabularx}

\par}
\bigskip
\centering
\captionsetup{singlelinecheck=false}
\caption{\textbf{Experience-injection ablation on RoboCasa365.} Per-task SR (\%) over 50 rollouts. }
\label{tab:injection}
{\fontsize{6.5}{7.5}\selectfont
\setlength{\tabcolsep}{1pt}
\begin{tabularx}{\linewidth}{@{}cc|>{\centering\arraybackslash}X>{\centering\arraybackslash}X>{\centering\arraybackslash}X>{\centering\arraybackslash}X>{\centering\arraybackslash}X|r@{}}
\toprule
\multicolumn{2}{@{}c|}{\textbf{Method}} & \multirow{2}{*}{{\fontsize{5.5}{6.5}\selectfont\bfseries\shortstack{TurnOn\\ElectricKettle}}} & \multirow{2}{*}{{\fontsize{5.5}{6.5}\selectfont\bfseries\shortstack{CloseToaster\\OvenDoor}}} & \multirow{2}{*}{{\fontsize{5.5}{6.5}\selectfont\bfseries\shortstack{TurnOn\\Microwave}}} & \multirow{2}{*}{{\fontsize{5.5}{6.5}\selectfont\bfseries\shortstack{CoffeeSetup\\Mug}}} & \multirow{2}{*}{{\fontsize{5.5}{6.5}\selectfont\bfseries\shortstack{OpenStand\\MixerHead}}} & \multirow{2}{*}{\textbf{Avg.}} \\
\cmidrule(lr){1-2}
\textbf{Token} & \textbf{Cross} &  &  &  &  &  &  \\
\midrule
 &  & 84.0 & 80.0 & \textbf{64.0} & 30.0 & 94.0 & 70.4 \\
$\checkmark$ &  & 92.0 & 77.0 & 62.0 & 49.0 & \textbf{96.0} & 75.2 \\
 & $\checkmark$ & 92.0 & 80.0 & \textbf{64.0} & 46.0 & 92.0 & 74.8 \\
\midrule
\rowcolor{tablerowgray}
$\checkmark$ & \multicolumn{1}{c|}{$\checkmark$} & \textbf{94.0} & \textbf{86.0} & \textbf{64.0} & \textbf{53.0} & \textbf{96.0} & \multicolumn{1}{|r@{}}{\textbf{78.6}} \\
\bottomrule
\end{tabularx}

\par}
\end{table}

\begin{figure}[t]
\centering
\includegraphics[width=\linewidth]{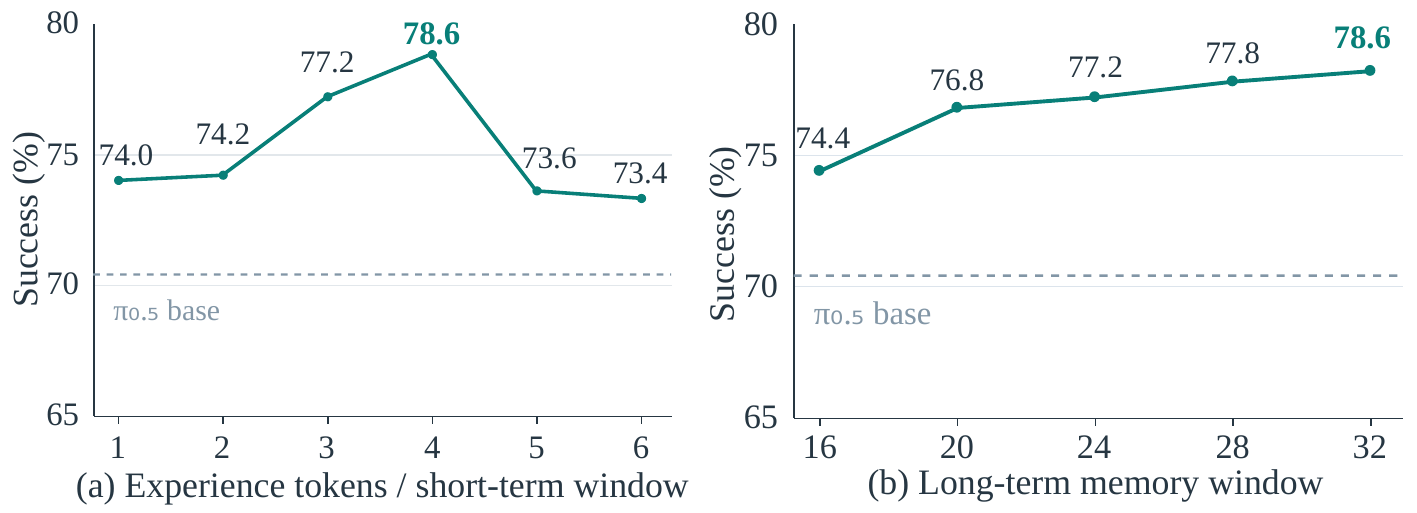}
\caption{\textbf{Memory-window ablations on RoboCasa365.} (a) Shared count $n$ for the short-term window, retrieval/readout slots, and experience tokens. (b) Long-term memory window. Dashed lines denote the base policy.}
\label{fig:token-count}
\end{figure}

\vspace{-6pt}
\subsection{Evaluation on Real-World Tasks}
\paragraph{Implementation Details.}
To evaluate \method in physical environments, we conduct experiments on an ARX~X5 bimanual platform with six joints and one gripper per arm.
We evaluate five manipulation tasks spanning short and long horizons.
All real-robot evaluations of \method use Phase 3 checkpoints.
For each task, we train on 250 demonstration episodes and conduct 20 evaluation trials with randomly initialized object positions.
We further evaluate demonstration-guided adaptation on seen tasks and unseen conditions.
Each demonstration-conditioned execution receives one successful demonstration.
Task splits and demonstration settings are detailed in the supplementary material.

\vspace{-12pt}
\paragraph{Results on real-world manipulation.}
As shown in Table~\ref{tab:real}, \method achieves the highest success rate on all five tasks, averaging 69.0\% compared with 60.0\% for \piBase and 43.0\% for Xiaomi-Robotics-1, the strongest external baseline by mean success.
Success increases from 30\% to 45\% on battery insertion, from 85\% to 95\% on block induction, and from 60\% to 70\% on clothes folding.
These tasks require tracking contact alignment, placement progress, or deformation, yet a single observation may not capture the effects of preceding actions.
We attribute these gains to structured memory storage and robust soft retrieval.
By organizing interactions into short- and long-term memory according to recency, \method accurately tracks task progress and retrieves the most relevant past experience to guide the current stage.

\begin{figure*}[t]
\centering
\includegraphics[width=\linewidth]{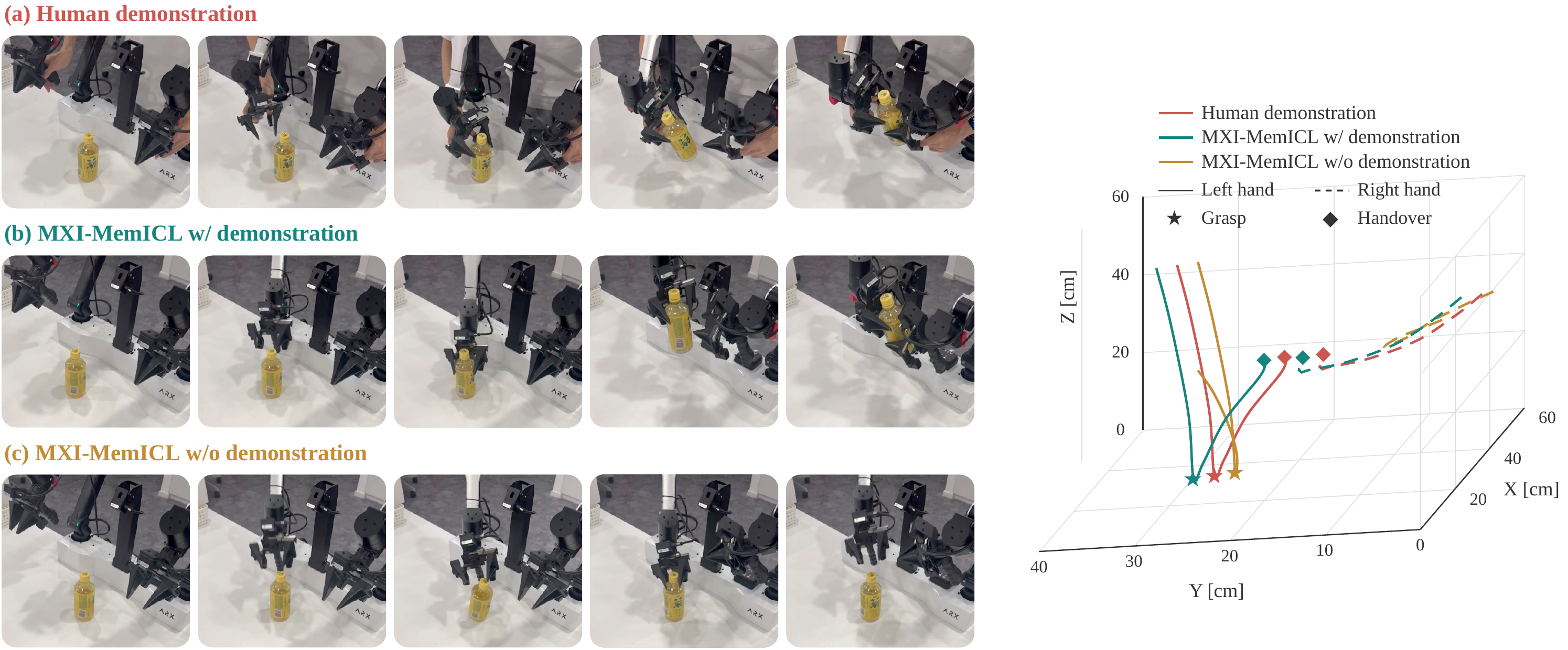}
\caption{\textbf{Demonstration-guided real-robot execution.} Rows show (a) a human demonstration, (b) \method with the demonstration, and (c) the \method without the demonstration. Frames progress from left to right. The right panel shows trajectories: solid/dashed lines denote left/right hands, and stars/diamonds mark grasp/handover events.}
\label{fig:qualitative}
\end{figure*}

\vspace{-12pt}
\paragraph{Demonstrations on seen tasks.}
On seen tasks, Figure~\ref{fig:real-icl}(a) reports mean success rates of 60.0\% for \piBase, 69.0\% for \method without demonstrations, and 74.0\% with a demonstration.
The demonstration-conditioned policy raises success on battery insertion from 45\% to 55\% and adds five points on block induction, clothes folding, and bottle-cap twisting, while water pouring remains at 90\%.
These demonstrations are encoded into varying numbers of memory embeddings according to their duration.
\method maintains substantial gains as task horizons increase, further demonstrating the robustness of our soft experience retrieval and adaptive experience injection strategies.

\vspace{-12pt}
\paragraph{Demonstrations on unseen tasks.}
On unseen tasks, Figure~\ref{fig:real-icl} (b) reports mean success rates of 21.7\%, 25.0\%, and 41.7\% for the same three models.
Demonstrations raise success from 50\% to 75\% on block induction, from 25\% to 40\% on clothes folding, and from 0\% to 10\% on bottle-cap twisting with different bottles.
The 16.7-point gain exceeds the 5.0-point gain on seen tasks, consistent with demonstrations providing experience relevant to the target conditions that online interactions alone may not yield.

\vspace{-6pt}
\subsection{Ablation Study}
We evaluate how memory timescales, experience injection, and memory windows affect success on RoboCasa365.
Memory-retention policies and evaluation repetitions are detailed in the supplementary material.

\vspace{-12pt}
\paragraph{Ablation on short-term and long-term memory.}
As shown in Table~\ref{tab:timescale}, the \piBase baseline achieves 70.4\% mean success.
Adding short-term memory alone raises success to 74.4\%, a 4.0-point gain that supports using recent interactions to track execution progress.
Long-term memory alone reaches 72.8\%, showing that experience retrieved from earlier interactions also provides useful action guidance.
Combining both yields 78.6\%, exceeding the baseline by 8.2 points and outperforming either component alone.
Together, recent context and retrieved history provide complementary guidance for action generation.

\vspace{-12pt}
\paragraph{Ablation on two-stream experience injection.}
Table~\ref{tab:injection} compares the two experience-injection routes.
Starting from the 70.4\% baseline, experience-token prefixes alone reach 75.2\%, supporting joint processing of experience, observations, and instructions at the input.
Gated cross-attention alone achieves 74.8\%, demonstrating the value of adaptively regulating experience injection within the backbone.
Combining both increases success to 78.6\%, a further 3.4-point gain over the stronger single route.
The combined gain highlights complementary experience guidance at the input and intermediate layers.

\paragraph{Ablation on short-term window and experience tokens.}
A shared hyperparameter $n$ determines the short-term memory window and the numbers of retrieval slots, readout slots, and injected experience tokens.
We therefore vary $n$ from 1 to 6 to examine how the amount of experience context affects manipulation performance.
Figure~\ref{fig:token-count} (a) shows that success rises from 74.0\% with one token to a peak of 78.6\% with four, both exceeding the 70.4\% baseline.
This gain supports providing richer experience context for action guidance.
Five and six tokens yield 73.6\% and 73.4\%, favoring the four-token configuration for concentrating useful experience rather than expanding context further.

\paragraph{Ablation on long-term memory window.}
We further investigate the impact of the long-term memory window on manipulation performance.
As shown in Figure~\ref{fig:token-count} (b), retaining 16 records yields 74.4\% success, 4.0 points above the 70.4\% baseline.
Expanding the window to 20, 24, 28, and 32 records progressively raises success to 76.8\%, 77.2\%, 77.8\%, and 78.6\%.
This trend supports retaining more completed interactions to provide intention-guided retrieval with a richer pool of experience for subsequent action guidance.
Together with the preceding ablation, these results favor a broader retrieval history paired with compact experience context for the policy.

\subsection{Analysis of In-Context Manipulation}
\label{sec:analysis}
Figure~\ref{fig:qualitative} (a) shows a human demonstration of twisting a bottle-cap.
Conditioned on this single demonstration, \method successfully completes the previously unseen task without parameter updates (Figure~\ref{fig:qualitative} (b)).
This shows that \method can retrieve experience from the demonstration to generate actions conditioned on the current observation rather than replaying the recorded action sequence.
The right-hand schematic contrasts the demonstrated and generated trajectories.
This execution illustrates how \method acquires and applies experience from a single demonstration to perform a new task.
In contrast, direct inference with \method fails to lift the bottle in Figure~\ref{fig:qualitative} (c), highlighting the value of demonstration experience for task adaptation.

\section{Conclusion}
In this work, we introduce \method, a framework for memory-guided in-context robot manipulation.
Specifically, \method encodes completed interactions into causal memory for closed-loop manipulation.
By anticipating next-stage intent, it retrieves relevant experience and adaptively injects it to guide subsequent actions.
By treating demonstrations as retrievable experience, \method enables robotic ICL without deployment-time parameter updates.
Experiments across diverse simulated and real-robot tasks demonstrate state-of-the-art success rates and effective adaptation to new tasks from a single demonstration.
\label{sec:body-end}

\section{AI-USE STATEMENT}
\label{sec:ai-use}
{\raggedright
We used generative AI tools to assist with translation, literature search, and reference checking.\par}

\clearpage
\setcounter{section}{0}
\setcounter{table}{0}
\setcounter{figure}{0}
\setcounter{equation}{0}
\renewcommand{\thetable}{S\arabic{table}}
\renewcommand{\thefigure}{S\arabic{figure}}
\renewcommand{\theequation}{S\arabic{equation}}
\renewcommand{\theHsection}{supp.\arabic{section}}
\renewcommand{\theHtable}{supp.\arabic{table}}
\renewcommand{\theHfigure}{supp.\arabic{figure}}
\renewcommand{\theHequation}{supp.\arabic{equation}}
\setcounter{topnumber}{2}
\setcounter{dbltopnumber}{2}
\makeatletter
\setlength{\@fptop}{0pt}
\setlength{\@dblfptop}{0pt}
\makeatother
\twocolumn[{\centering\Large\bfseries Supplementary Material\par\vspace{12pt}}]
\begingroup
\setlength{\stripsep}{0pt}
\begin{strip}
\centering
\includegraphics[width=0.94\textwidth]{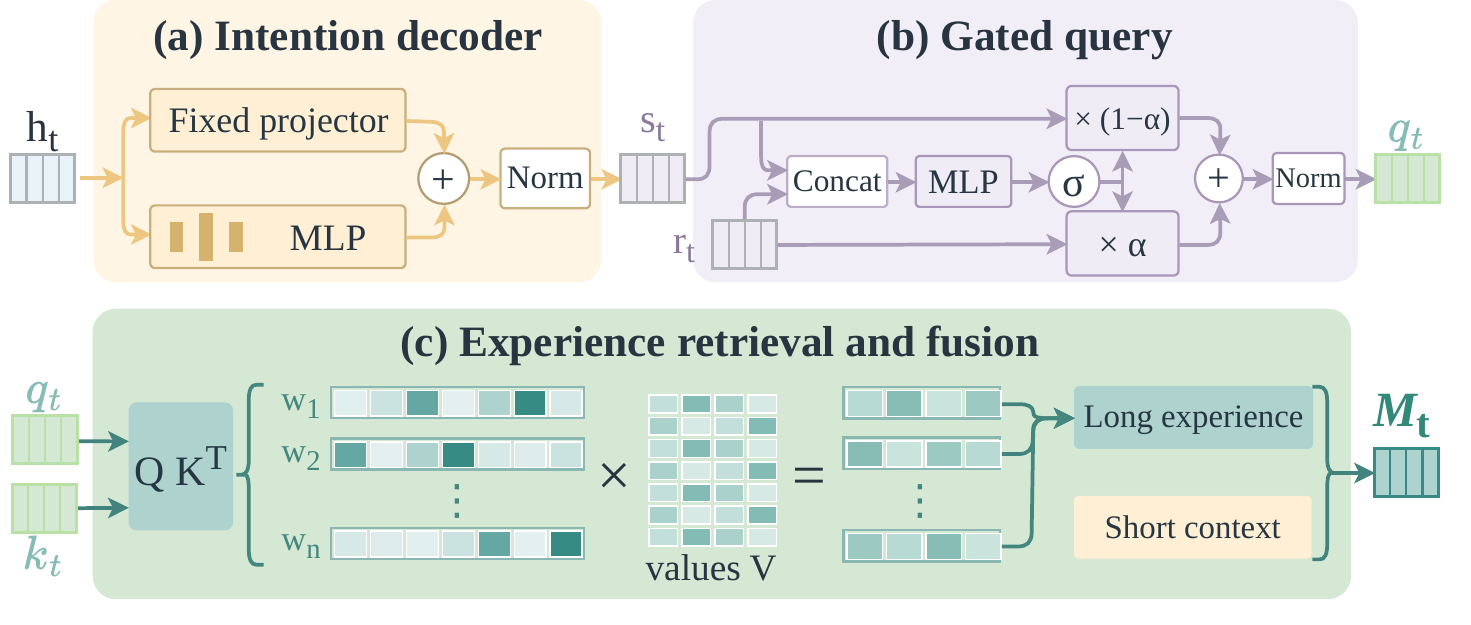}
\captionof{figure}{\textbf{Module details for main-paper Figure~2.}
(a) \textbf{The intention decoder and intention predictor use the same residual MLP design with separate weights.}
The decoder adds its residual to a fixed projection of $h_t$, and the predictor adds its residual to $q_t$.
(b) A learned gate combines the stage embedding $s_t$ and prior intent $r_t$ into the query $q_t$.
(c) Each slot normalizes its query--key scores over available records.
The resulting weights pool memory values, and the retrieved experience is fused with short-term context to form the experience tokens $M_t$.}
\label{fig:supp-method-modules}
\par\vspace{10pt}
\end{strip}
\endgroup

\section{Implementation and Training Details}
\label{sec:supp-method}

\paragraph{Policy training and deployment.}
Unless otherwise specified, we train the memory components for 8k updates in Phase 2 and 4k in Phase 3 on eight NVIDIA PRO 6000 GPUs with a batch size of 64, following the main-paper configuration.
We use Phase 3 checkpoints to evaluate \method both in simulation and on real robots.
At each decision, the policy predicts 50 controls.
The controller executes the first 16 and adds the completed interaction to online memory.

\subsection{Representation parameterization}
The visual--action causal encoder uses GRU temporal mixers and same-timestep cross-stream attention.
A memory window contains four transitions, each spanning four executed controls.
The recurrent hidden state has width 256, while normalized effect and retrieval representations have width 128.
State persists within an episode, and a window enters memory after all 16 controls have been executed.
Pooling over valid observation and instruction positions produces the decoder input $h_t$ in Figure~\ref{fig:supp-method-modules}(a).
The intention decoder has hidden and output widths of 256 and 128, and combines a fixed semantic anchor with a learned residual:
\begin{align}
 u_t&=\norm{R\operatorname{LN}(h_t)},\nonumber\\
 s_t&=D_\phi(h_t)=\norm{u_t+f_{\rm cur}(h_t)}.
\end{align}
The fixed projection $R$ anchors the normalized backbone feature, while $f_{\rm cur}$ adapts the stage embedding.
For each key--memory pair, the implementation caches the underlying features $(h_t,z_t,z_{t+1},e_t,a_t^{\rm sum})$: the backbone feature, the interaction states before and after execution, the visual effect, and an action summary of width $4d_a$.
MLP projections construct the retrieval key and memory value:
\begin{align}
 k_t&=\norm{u_t+f_k([u_t,z_t])},\\
 m_t&=f_v([z_t,z_{t+1},e_t,a_t^{\rm sum}]).
\end{align}
Projecting cached features allows the key and memory mappings to remain trainable.
Record indices preserve source identity and temporal order.

\subsection{Retrieval and experience fusion}
A shared hyperparameter $n=4$ sets the short-term window, retrieval and readout slots, and experience-token count unless otherwise specified.
The gated query mapping $G_\phi$ uses a sigmoid gate to combine $s_t$ and $r_t$.
Each slot adds a learned offset $d_r$, allowing projected queries to match different records through cosine similarity:
\begin{equation}
 \rho_{r,i}=\cos(W_q\norm{q_t+d_r},W_k k_i).
\end{equation}
With temperature $\tau=0.1$, record availability $b_{r,i}$ modulates the soft retrieval weights:
\begin{equation}
 w_{r,i}=\frac{b_{r,i}\exp(\rho_{r,i}/\tau)}
 {\sum_{j\in\mathcal C_t}b_{r,j}\exp(\rho_{r,j}/\tau)}.
\end{equation}
The candidate set $\mathcal C_t$ combines long-term online memory with the read-only demonstration bank when available, while short-term records enter a separate fusion branch.
Valid records start with $b_{1,i}=1$.
Updating their availability after each slot discourages repeated retrieval:
\begin{equation}
 b_{r+1,i}=\max(10^{-6},b_{r,i}(1-w_{r,i})).
\end{equation}
Invalid records are masked before normalization.
The two attention branches in the main paper's experience fusion step use $n$ learned queries $Q_t=Q+W_Iq_t$ to produce
\begin{align}
 U_t^S&=\operatorname{Attn}_S(Q_t,V_t^{\rm short}),\nonumber\\
 U_t^L&=\operatorname{Attn}_L(Q_t,\widetilde{\mathcal M}_t),\nonumber\\
 M_t&=W_O\operatorname{GELU}\!\left(\operatorname{LN}(W_F[U_t^S,U_t^L])\right).
\end{align}
$V_t^{\rm short}$ contains the memory values in $\mathcal B_t^{\rm short}$, and each attention branch uses a learned null value for empty inputs.
In the 18-layer backbone, gated cross-attention operates at layers $0,3,6,9,12,15$, after the self-attention residual and before the feed-forward block.
In the cross-attention-only ablation, $\bar h_t$ pools valid observation and instruction positions.

\paragraph{Retention and resets.}
All configurations use FIFO retention during both training and inference, removing the oldest online record when capacity is reached.
Each evaluation rollout starts with an empty online bank and resets recurrent intent and interaction-encoder state.
When supplied, a successful demonstration is encoded separately before execution and remains read-only throughout the rollout.

\subsection{Training objectives and sequence handling}

\paragraph{Phase 1.}
Interaction encoding combines action and visual prediction with task, phase, and effect supervision:
\begin{equation}
\begin{split}
 \mathcal L_{\rm I}={}&\mathcal L_{\rm act}+\mathcal L_{\rm vis}
 +0.2\mathcal L_{\rm task}\\
 &+0.1\mathcal L_{\rm phase}+0.25\mathcal L_{\rm effect}.
\end{split}
\label{eq:encoder-loss}
\end{equation}
Masked Smooth L1 loss supervises action prediction, and squared error aligns predicted visual features with a stop-gradient next-frame target.
Multi-positive contrast groups examples sharing a task identity.
Phase regularization penalizes latent increments that fail to advance along the trajectory's first-to-last direction.
Effect supervision combines contrastive matching to visual change, action-summary prediction, predicted/observed effect alignment, and variance/covariance regularization.
The observed-effect branch takes visual features, and the memory-value projection additionally receives executed actions and recurrent states.

\paragraph{Phases 2 and 3.}
The simulation configurations use $\lambda_I=0.1$ and $\lambda_A=0.01$ for the main-paper intent and anchor losses.
The flow-matching term perturbs the action chunk $A_t$ with Gaussian noise $\epsilon$ to form $X_{t,\sigma}=(1-\sigma)A_t+\sigma\epsilon$:
\begin{equation}
 \mathcal L_{\rm FM}=\mathbb E\!\left[
 \|v_{\theta,\phi}(X_{t,\sigma},\sigma\mid x_t,M_t)-(\epsilon-A_t)\|_\Omega^2
 \right].
 \label{eq:flow}
\end{equation}
The supervision mask $\Omega$ selects valid action dimensions and excludes padding and the known action prefix in RTC-enabled ARX~X5 training.
Phase 3 applies the same objective to target execution.
Training windows begin with a burn-in segment that reconstructs recurrent intent.
We detach the resulting state, then propagate gradients through the intent predictions across consecutive supervised decisions.
Demonstration and execution recordings are grouped by source, with disjoint groups across training and validation.

\section{Simulation Benchmarks}
\label{sec:supp-protocols}
We evaluate each simulation experiment at least twice, with per-task rollout counts below referring to each run.
All simulation evaluations are conducted without test-time demonstrations.
We report per-task success rates and their equally weighted mean.

\subsection{RoboTwin 2.0}
\label{sec:supp-robotwin}
\begin{table*}[t]
\centering
\caption{\textbf{Performance comparison on additional RoboTwin 2.0 tasks.}
Per-task SR (\%) over 100 rollouts per run under the Hard setting.
These 12 selected tasks show positive gains over our reproduced $\pi_{0.5}$ beyond main-paper Table~1.
$\dagger$ denotes our reproduced result.
Other baselines are from the leaderboard.}
\label{tab:robotwin-additional}
{\small
\setlength{\tabcolsep}{2.5pt}
\begin{tabularx}{\linewidth}{@{}l|>{\hsize=1.05\hsize\linewidth=\hsize\centering\arraybackslash}X>{\hsize=1.05\hsize\linewidth=\hsize\centering\arraybackslash}X>{\hsize=0.65\hsize\linewidth=\hsize\centering\arraybackslash}X>{\hsize=0.95\hsize\linewidth=\hsize\centering\arraybackslash}X>{\hsize=1.3\hsize\linewidth=\hsize\centering\arraybackslash}X>{\hsize=1.05\hsize\linewidth=\hsize\centering\arraybackslash}X>{\hsize=0.85\hsize\linewidth=\hsize\centering\arraybackslash}X>{\hsize=1.1\hsize\linewidth=\hsize\centering\arraybackslash}X@{}}
\toprule
\textbf{Task} & \shortstack{Fast-WAM\\\cite{fastwam}} & \shortstack{Discrete\\Forcing~\cite{discreteforcing}} & $\pi_0$~\cite{pi0} & \shortstack{X-VLA\\\cite{xvla}} & \shortstack{Xiaomi-\\Robotics-0~\cite{xiaomi0}} & \shortstack{EventVLA\\\cite{eventvla}} & $\pi_{0.5}$~\cite{pi05}$^{\dagger}$ & \textbf{\method} \\
\midrule
Rank Blocks (Size) & 0.0 & 2.0 & 1.0 & \textbf{25.0} & 1.0 & 1.0 & 22.0 & \textbf{25.0} \\
Handover Mic & 0.0 & 2.0 & 13.0 & 1.0 & 6.0 & 5.0 & 46.0 & \textbf{52.0} \\
Move Pill Bottle to Pad & 0.0 & 0.0 & 1.0 & 3.0 & 3.0 & 3.0 & 11.0 & \textbf{16.0} \\
Move Stapler to Pad & 0.0 & 0.0 & 2.0 & 0.0 & 0.0 & 2.0 & 10.0 & \textbf{13.0} \\
Pick Diverse Bottles & 0.0 & 10.0 & 6.0 & 30.0 & 12.0 & 9.0 & 46.0 & \textbf{49.0} \\
Place A2B Left & 0.0 & 10.0 & 1.0 & 15.0 & 6.0 & 20.0 & 39.0 & \textbf{40.0} \\
Place Burger with Fries & 0.0 & 2.0 & 4.0 & 35.0 & 27.0 & 7.0 & 83.0 & \textbf{88.0} \\
Place Empty Cup & 0.0 & 6.0 & 11.0 & 16.0 & 12.0 & 18.0 & 57.0 & \textbf{63.0} \\
Place Object on Scale & 0.0 & 2.0 & 0.0 & 6.0 & 2.0 & 24.0 & 33.0 & \textbf{38.0} \\
Place Object on Stand & 0.0 & 6.0 & 11.0 & 43.0 & 24.0 & 32.0 & 51.0 & \textbf{52.0} \\
Shake Bottle & 10.0 & 66.0 & 60.0 & 76.0 & 61.0 & 34.0 & 96.0 & \textbf{99.0} \\
Shake Bottle Horizontally & 9.0 & 70.0 & 51.0 & 78.0 & 62.0 & 37.0 & 98.0 & \textbf{100.0} \\
\midrule
Average & 1.6 & 14.7 & 13.4 & 27.3 & 18.0 & 16.0 & 49.3 & \textbf{52.9} \\
\bottomrule
\end{tabularx}

\par}
\end{table*}

We train our RoboTwin models using 50 demonstrations per task, matching the training-data setting of the leaderboard baselines.
The reproduced $\pi_{0.5}$ base is fine-tuned for 30k updates on eight NVIDIA PRO 6000 GPUs with a global batch size of 128.
We use AdamW with $(\beta_1,\beta_2)=(0.9,0.95)$ and weight decay of 0.01.
The learning rate increases linearly to $2.5\times10^{-5}$ over 1,000 warm-up updates, then follows a cosine schedule to $2.5\times10^{-6}$.
Gradients are clipped to a global norm of 1.0.
Building on this base model, we train \method for 4k updates in Phase 1, followed by the shared Phase 2 and 3 schedule of 8k and 4k updates at batch size 64.
The reported $\pi_0$ policies are trained separately for each task, while the other leaderboard policies use joint training.
The ten tasks in main-paper Table~1 and twelve additional tasks in Table~\ref{tab:robotwin-additional} use the same checkpoints, with 100 rollouts per task per run under the Hard setting.
The external baselines retain the model identities and training settings of main-paper Table~1, using Hard scores from the \href{https://robotwin-platform.github.io/leaderboard\#ranking}{RoboTwin leaderboard}.
Across these tasks, mean success rises from 49.3\% to 52.9\%.
Handover Mic and Place Empty Cup each gain 6 percentage points, while Move Pill Bottle to Pad, Place Burger with Fries, and Place Object on Scale each gain 5 points.
The gains span handover, transport, and placement, extending the evidence for experience-conditioned control beyond the tasks in the main table.

\subsection{RoboCasa365}
\label{sec:supp-extension10}
Each of the five RoboCasa365 tasks in the main paper is evaluated over at least two runs of 50 rollouts each, totaling at least 100 rollouts per task.
Every run reuses the same 50 fixed scenes with seeds 210000--210049, and online memory is reset before each rollout.
Budgets are 450 controls, or 600 for Coffee Setup Mug.
The $\pi_{0.5}$ family uses ten flow-matching steps with the action-chunk settings in Section~\ref{sec:supp-method}.
Other architectures retain their training recipes.
Beyond the five tasks in the main paper, we evaluate eight additional RoboCasa365 tasks involving articulated fixtures, object transport, and multi-step kitchen routines.
Both policies are evaluated over at least two runs on the same 400 fixed target-split scenes, with 50 rollouts per task per run, totaling at least 800 rollouts per policy and 100 per task.

\begin{table}[t]
\centering
\caption{\textbf{Performance comparison on eight additional RoboCasa365 tasks.}
Per-task SR (\%) over 50 matched scenes per run, with one rollout per scene.}
\label{tab:extension10}
{\small\setlength{\tabcolsep}{4pt}
\begin{tabularx}{\linewidth}{@{}>{\raggedright\arraybackslash}X|rr@{}}
\toprule
\textbf{Task} & $\pi_{0.5}$ & \textbf{\method} \\
\midrule
Get Toasted Bread & 34.0 & \textbf{36.0} \\
Load Dishwasher & 30.0 & \textbf{44.0} \\
Open Cabinet & 82.0 & \textbf{84.0} \\
Open Drawer & \textbf{78.0} & \textbf{78.0} \\
Prepare Coffee & \textbf{26.0} & 24.0 \\
Slide Dishwasher Rack & 82.0 & \textbf{86.0} \\
Steam In Microwave & 16.0 & \textbf{20.0} \\
Turn On Sink Faucet & 54.0 & \textbf{62.0} \\
\midrule
\multicolumn{1}{@{}l|}{Average} & 50.3 & \textbf{54.3} \\
\bottomrule
\end{tabularx}

\par}
\end{table}

\begin{table*}[t]
\centering
\caption{\textbf{Performance comparison on LIBERO.}
Per-suite SR (\%).
$\dagger$ denotes our reproduced result.}
\label{tab:libero-comparison}
\fontsize{8}{9.5}\selectfont
\begin{tabularx}{\textwidth}{@{}l|*{4}{>{\centering\arraybackslash}X}|>{\centering\arraybackslash}X@{}}
\toprule
\textbf{Method} & \textbf{Spatial} & \textbf{Object} & \textbf{Goal} & \textbf{LIBERO-10} & \textbf{Avg.} \\
\midrule
Diffusion Policy~\cite{dp} & 78.3 & 92.5 & 68.3 & 50.5 & 72.4 \\
GR00T N1~\cite{grootn1} & 94.4 & 97.6 & 93.0 & 90.6 & 93.9 \\
OpenVLA~\cite{openvla} & 84.7 & 88.4 & 79.2 & 53.7 & 76.5 \\
OpenVLA-OFT~\cite{openvlaoft} & 97.6 & 98.4 & 97.9 & 94.5 & 97.1 \\
MemoryVLA~\cite{memoryvla} & 98.4 & 98.4 & 96.4 & 93.4 & 96.7 \\
Unified VLA~\cite{unifiedvla} & 95.4 & 98.8 & 93.6 & 94.0 & 95.5 \\
InternVLA-M1~\cite{internvlam1} & 98.0 & 99.0 & 93.8 & 92.6 & 95.9 \\
$\pi_0$-FAST~\cite{fast} & 96.4 & 96.8 & 88.6 & 60.2 & 85.5 \\
$\pi_0$~\cite{pi0} & 96.8 & 98.8 & 95.8 & 85.2 & 94.2 \\
BehaviorVLA~\cite{behaviorvla} & \textbf{99.2} & \textbf{99.4} & \textbf{98.8} & 94.6 & 98.0 \\
\midrule
$\pi_{0.5}^{\dagger}$ ~\cite{pi05} & 97.2 & 99.2 & 97.2 & 95.6 & 97.3 \\
\rowcolor{tablerowgray}
\multicolumn{1}{@{}l|}{\cellcolor{tablerowgray}\textbf{\method{} (Ours)}} & 98.0 & \textbf{99.4} & 98.0 & \textbf{97.0} & \textbf{98.1} \\
\bottomrule
\end{tabularx}

\end{table*}

\paragraph{Training and evaluation.}
Training uses 4000 complete demonstrations from the original eight-task set.
The base policy receives 30k updates, followed by 5k for the causal encoder in Phase 1 and the shared Phase 2 and 3 schedule in Section~\ref{sec:supp-method}.
Evaluation retains all 12 action dimensions for mobile manipulation and otherwise follows the RoboCasa365 action-horizon and flow-matching settings.
Scene seeds 210000--210049 and task-specific control budgets are shared across policies.
Control budgets range from 450 to 1,050 steps for atomic operations and from 1,800 to 3,000 for multi-step tasks.

\paragraph{Results.}
Across the eight additional tasks, \method reaches 54.3\% mean success compared with 50.3\% for the base policy (Table~\ref{tab:extension10}).
The largest gains occur in Load Dishwasher and Turn On Sink Faucet, where success rises from 30.0\% to 44.0\% and from 54.0\% to 62.0\%, respectively.
Six of eight tasks improve, and Open Drawer is unchanged.
\subsection{LIBERO}
\label{sec:supp-libero}

Table~\ref{tab:libero-comparison} reports results on LIBERO~\cite{libero}.
We compare the Phase 3 \method policy with the 30k-update base.
External models retain their published protocols.

\paragraph{Results and analysis.}
Table~\ref{tab:libero-comparison} shows the largest gain over our reproduced base on LIBERO-10: 97.0\% versus 95.6\%.
This is the highest LIBERO-10 score among the listed methods, including BehaviorVLA at 94.6\%~\cite{behaviorvla}.
Compared with 0.8-point gains on Spatial and Goal, the larger long-horizon gain supports retaining execution history for sustained task completion.
\method achieves 98.1\% mean success on LIBERO.
These results demonstrate consistent gains over our reproduced \piBase baseline.

\begin{table}[!tb]
\centering
\caption{\textbf{Intention and retrieval controls on RoboCasa365.}
Mean SR (\%) over two runs on the same 250 scenes.}
\label{tab:intent-controls}
{\small\setlength{\tabcolsep}{4pt}\begin{tabular}{@{}p{\dimexpr0.85\linewidth-40pt-2\tabcolsep-\arrayrulewidth\relax}|>{\centering\arraybackslash}p{40pt}@{}}
\toprule
\textbf{Variant} & \textbf{Avg.} \\

\midrule
\piBase{} & 70.4 \\
Current stage ($q_t=s_t$) & 72.2 \\
Current-feature top-$k$ & 73.6 \\
No intent loss & 74.0 \\
\midrule
\rowcolor{tablerowgray}
\method{} (ours) & \textbf{78.6} \\
\bottomrule
\end{tabular}
}
\end{table}
\begin{table*}[!t]
\centering
\fontsize{7}{8.5}\selectfont
\setlength{\tabcolsep}{2pt}
\caption{\textbf{Latency and execution time on RoboCasa365.}
Per-task completion times (s), inference latency (ms), and SR (\%) on five selected tasks.
Completion times average successful rollouts and exclude scene setup.}
\label{tab:latency-robocasa}
\begin{tabularx}{\linewidth}{@{}p{101pt}|*{9}{>{\centering\arraybackslash}X}@{}}
\toprule
\multicolumn{1}{@{}l|}{\textbf{Model}} & \multicolumn{6}{c}{\textbf{Mean task completion time (s)} $\downarrow$} & \multicolumn{2}{c}{\textbf{Inference (ms)} $\downarrow$} & \textbf{SR (\%)} $\uparrow$ \\
\cmidrule(lr){2-7}\cmidrule(lr){8-9}
 & \shortstack{TurnOn\\ElectricKettle} & \shortstack{CloseToaster\\OvenDoor} & \shortstack{TurnOn\\Microwave} & \shortstack{CoffeeSetup\\Mug} & \shortstack{OpenStand\\MixerHead} & \textbf{Avg. (5)} & Mean & P95 & Avg. (5) \\
\midrule
Xiaomi-Robotics-1~\cite{xiaomi1} & 7.7 & 10.5 & 7.6 & 10.6 & 5.6 & 8.4 & 202.4 & 209.6 & 70.8 \\
Fast-WAM~\cite{fastwam} & 13.2 & 23.9 & 14.4 & 22.1 & 10.7 & 16.9 & 772.2 & 735.6 & 56.4 \\
Cosmos3-Nano~\cite{cosmos3} & 27.1 & 30.8 & 42.8 & 42.3 & 21.3 & 32.9 & 1458.2 & 1481.1 & 56.0 \\
$\tau_0$-WM~\cite{tau0wm} & 17.6 & 22.0 & 17.9 & 26.5 & 18.2 & 20.5 & 467.6 & 502.2 & 45.2 \\
$\pi_{0.5}$~\cite{pi05} & 7.1 & 8.0 & 6.6 & 10.2 & 4.9 & 7.4 & 80.0 & 84.9 & 70.4 \\
\midrule
\rowcolor{tablerowgray}
\multicolumn{1}{@{}p{101pt}|}{\shortstack[l]{\method{} (Ours)}} & 8.6 & 10.4 & 9.2 & 13.3 & 6.5 & 9.6 & 226.8 & 342.6 & 78.6 \\
\bottomrule
\end{tabularx}

\par\medskip
\begin{minipage}[t]{\columnwidth}
\vspace{0pt}
\centering
\small
\setlength{\tabcolsep}{2pt}
\caption{\textbf{Inference latency on real-robot tasks.}
Mean and P95 latency (ms), with SR (\%) from main-paper Table~3.
$\dagger$ denotes action-RPC upper bounds.}
\label{tab:latency-real}
\begin{tabularx}{\linewidth}{@{}l|*{3}{>{\centering\arraybackslash}X}@{}}
\toprule
\multicolumn{1}{@{}l|}{\textbf{Model}} & \multicolumn{2}{c}{\textbf{Inference (ms)} $\downarrow$} & \textbf{SR (\%)} $\uparrow$ \\
\cmidrule(lr){2-3}
 & Mean & P95 & Avg. (5) \\
\midrule
Cosmos3-Nano~\cite{cosmos3} & 706.2 & 767.9 & 17.0 \\
Fast-WAM~\cite{fastwam} & 210.4 & 232.8 & 38.0 \\
$\tau_0$-WM~\cite{tau0wm} & 390.8 & 435.1 & 37.0 \\
Xiaomi-Robotics-1~\cite{xiaomi1} & 369.9 & 389.5 & 43.0 \\
$\pi_{0.5}$~\cite{pi05} & 64.5 & 79.9 & 60.0 \\
\midrule
\rowcolor{tablerowgray}
\multicolumn{1}{@{}l|}{\method{} (Ours)} & $\leq 111.6^{\dagger}$ & $\leq 134.7^{\dagger}$ & 69.0 \\
\bottomrule
\end{tabularx}

\end{minipage}\hfill
\begin{minipage}[t]{\columnwidth}
\vspace{0pt}
\centering
\includegraphics[width=\linewidth]{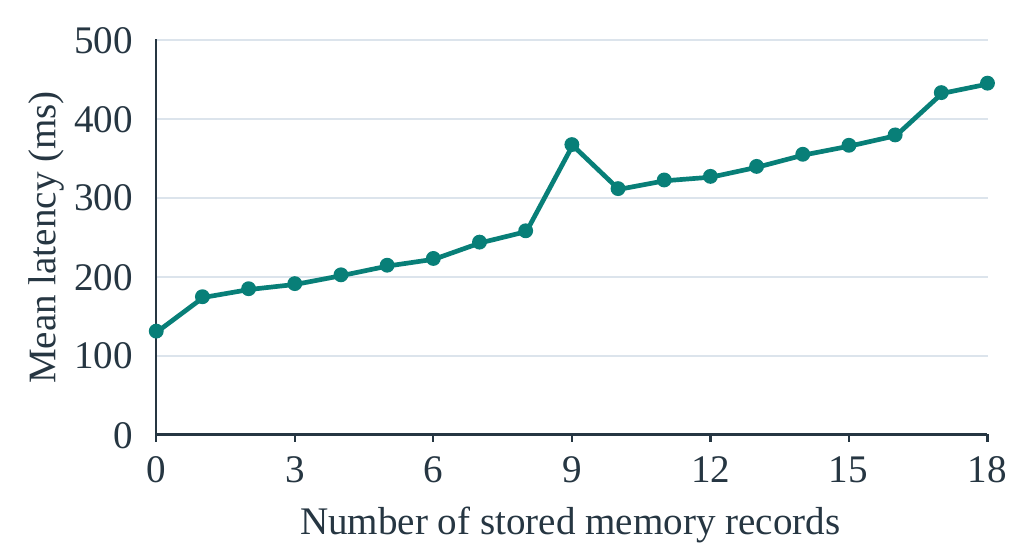}
\captionof{figure}{\textbf{Inference latency versus memory size.}
Each point shows the mean per-query latency at a given memory size across successful rollouts on the five RoboCasa365 tasks.}
\label{fig:supp-memory-latency}
\end{minipage}
\end{table*}

\section{Real-Robot Evaluation Protocols}
\label{sec:supp-real-icl}
The ARX~X5 platform uses one external-view camera and two wrist-mounted cameras, with images resized to $224\times224$ before model input.
Each task uses 250 training demonstrations and 20 evaluation trials with randomly initialized object positions.
The main real-robot comparison uses the Phase 3 controller without test-time demonstrations.
Main-paper Figure 3b uses the held-out conditions below.
The training phases follow Section 3.5 of the main paper and exclude base-policy preparation.
Real-robot experiments follow the shared training schedule of 8k updates in Phase 2 and 4k in Phase 3.

\paragraph{Block induction.}
The color and row splits exclude red targets and first-row targets, respectively.
Each split retains complete trajectories from 18 target conditions and holds out nine target conditions.
The base policy and Phase 1 causal encoder use the corresponding training subset for 20k and 5k updates, respectively.
Phase 3 uses demonstration--execution pairs drawn from the corresponding training subset.
Both sides of each pair respect the held-out targets, yielding 2,436 pairs for the color split and 2,507 for the row split.
The unseen block result in the main paper aggregates 20 evaluation trials: ten under the red-target exclusion split and ten under the first-row exclusion split.
The source recordings of the evaluation demonstrations are excluded from every training phase.

\paragraph{Clothes folding.}
We partition the data by the initial orientation of the garment collar.
Training covers forward-, backward-, and left-facing collars with angular deviations of up to $\pm30^\circ$, while right-facing configurations are held out for test-time demonstrations and evaluation.
All training phases follow this orientation-based split, and normalization statistics are computed from training demonstrations.
The base policy and Phase 1 causal encoder receive 25k and 5k updates, respectively.
Phase 3 pairs distinct successful training recordings with matching collection batches and task instructions.

\paragraph{Bottle-cap twisting.}
Evaluation uses different bottles from training, with a successful demonstration of the test bottle supplied at inference.

\section{Ablation Studies}

\paragraph{Evaluation and memory retention.}
All ablation configurations are evaluated twice on the same 250 scenes across five tasks, with 50 rollouts per task per run.
The single-timescale variants include an action-prior branch.
The full method and all ablation variants use FIFO memory retention during both training and inference.
For the first token-only execution, a paired McNemar comparison against the base yields $p=0.0919$.

\paragraph{Ablation on intention-guided retrieval.}
To test whether anticipating the next stage helps select experience for action, we compare three retrieval variants on five RoboCasa365 tasks (Table~\ref{tab:intent-controls}).
The current-stage control bypasses the gated query during both training and inference by setting $q_t=s_t$, excluding predicted intent $r_t$ from the query while retaining learned keys, soft retrieval, intent supervision, and two-stream experience injection.
The current-feature top-$k$ control replaces intention-guided soft retrieval with cosine matching between current and stored backbone features $h$, selecting up to $n=4$ distinct records and using $q_t=s_t$ for fusion while retaining causal memory values, short-term context, and both injection paths.
The no-intent-loss control sets $\lambda_I=0$ during training to remove prediction supervision from the projected next-step backbone feature, while retaining the intention predictor, gated query, anchor loss, and action-generation objective.
Current-feature retrieval raises success from 70.4\% for \piBase{} to 73.6\%, showing the benefit of matching experience to the current observation.
Querying memory with the current stage embedding alone yields 72.2\% success.
Incorporating the intent $r_t$ predicted at the preceding decision raises success to 78.6\%.
Removing the intent loss yields 74.0\% success, 4.6 points below the full method.
Together, these comparisons show the value of both learning to anticipate the next stage and using that prediction to retrieve experience for the upcoming action.

\section{Additional Analyses}
\begin{figure*}[!t]
\centering
\includegraphics[width=\linewidth]{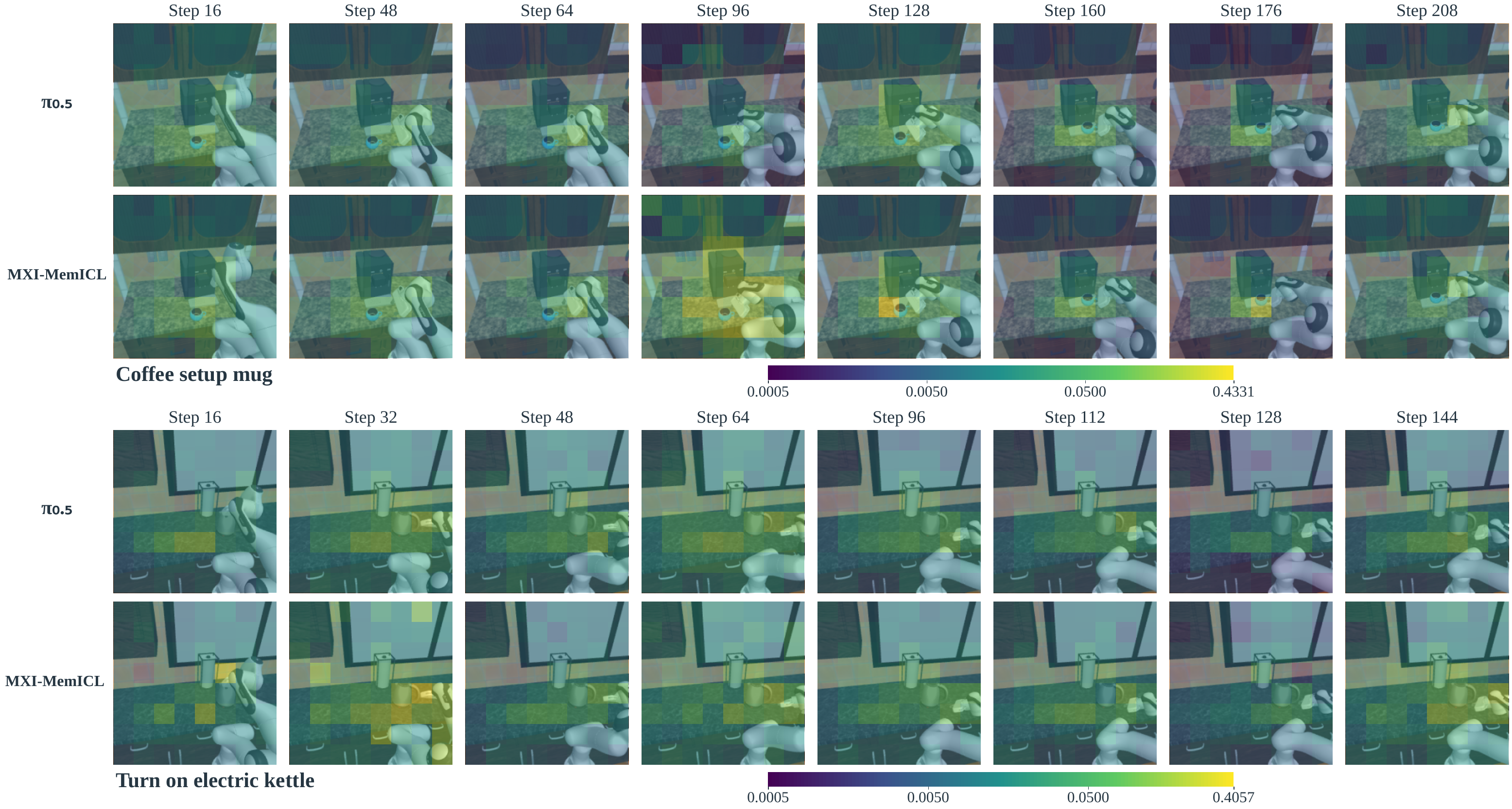}
\caption{\textbf{Visual sensitivity of action generation.}
Paired rows compare $\pi_{0.5}$ and \method on identical observations from Coffee Setup Mug and Turn On Electric Kettle.
Warmer colors indicate larger changes in predicted actions under image-patch occlusion.
Each task uses its own logarithmic scale, shared across both policies and all eight frames.}
\label{fig:supp-visual-sensitivity}
\end{figure*}

\subsection{Inference latency and task execution time}
\label{sec:supp-latency}
Following OpenVLA-OFT~\cite{openvlaoft} and RTC~\cite{rtc2025}, we distinguish action-chunk inference latency from end-to-end task completion time (Tables~\ref{tab:latency-robocasa} and~\ref{tab:latency-real}).

\paragraph{Simulation protocol.}
Task time runs from the end of snapshot restoration to official success, including inference waits and environment stepping but excluding setup and warmup.
Means use successful rollouts with equal task weights.
Each run shares one NVIDIA PRO 6000 between policy and simulator.
Concurrent runs use separate GPUs but share host CPUs.
Models retain their checkpoint and decoding settings at 20\,Hz simulation.
Server latency includes memory processing but excludes websocket transport.
We give each task equal weight when aggregating mean and P95 latency.

\paragraph{From inference to execution.}
Table~\ref{tab:latency-robocasa} reports a mean inference latency of 226.8\,ms for \method versus 80.0\,ms for \piBase, while mean completion time on successful rollouts increases from 7.4 to 9.6\,s.
The roughly $2.8\times$ per-call latency translates into a $1.3\times$ completion time because each predicted action chunk supplies multiple controls and total task time also includes environment stepping.
Experience retrieval, fusion, and injection add computation to each decision, but this cost is amortized across the controls in an action chunk.
Together with the success-rate increase from 70.4\% to 78.6\%, these results quantify the execution overhead associated with more reliable manipulation.
These results indicate that \method delivers substantial gains in manipulation success with inference overhead that remains manageable at the task level.

\paragraph{Memory-dependent latency.}
In Figure~\ref{fig:supp-memory-latency}, mean query latency across successful rollouts rises from 130.6\,ms with no stored records to 257.5\,ms with eight records and 378.8\,ms with sixteen.
As the bank grows, soft retrieval processes more candidate records even though the number of experience tokens injected into the backbone remains fixed.
This trend helps explain the P95 of 342.6\,ms relative to the 226.8\,ms mean, as decisions carry different history lengths and corresponding computational loads.
The fixed experience-token count limits context overhead in the backbone, while memory capacity controls the extent and processing cost of retrievable history.

\paragraph{Real-robot inference.}
\method action-RPC timings include communication and exclude separate memory-update calls, so the table marks them as upper bounds on action computation.
Table~\ref{tab:latency-real} places \method's mean and P95 action-computation latency at no more than 111.6 and 134.7\,ms, compared with 64.5 and 79.9\,ms for \piBase.
The real-robot success rate increases from 60.0\% to 69.0\%, pairing the additional action-generation computation with more reliable manipulation.

\subsection{Visual sensitivity of action generation}
\label{sec:supp-visual-sensitivity}
Figure~\ref{fig:supp-visual-sensitivity} compares the frozen $\pi_{0.5}$ base and \method at eight sampled decisions per task.
Coffee Setup Mug and Turn On Electric Kettle provide examples of object transport and appliance control.

\paragraph{Occlusion protocol.}
We divide each observation into an $8\times8$ grid, replace one patch at a time with the image's channel-wise mean, and measure the root-mean-square change in the normalized predicted action prefix.
For each paired comparison, robot state, instruction, random noise, and the pre-decision history remain fixed.
\method recomputes its query and retrieved experience from the perturbed observation.
All frames and both policies share a logarithmic scale within each task, with separate scales for the two tasks.
Colors show changes in action predictions under visual occlusion.

\paragraph{Stage-dependent visual sensitivity.}
Figure~\ref{fig:supp-visual-sensitivity} shows that the two policies respond differently to visual occlusion at different execution stages, even when given identical observations.
In Coffee Setup Mug, \method exhibits a stronger response around the mug and gripper at step 96, where the base policy is less sensitive.
By step 128, both policies respond strongly to occlusion near the mug, indicating that the difference depends on the execution stage.
In Turn On Electric Kettle, \method responds strongly near the gripper and kettle at step 32, followed by a weaker response at step 48.
These comparisons show that experience-conditioned action generation varies its dependence on local visual information over the interaction, consistent with using execution context to adapt the current action.

\FloatBarrier
\end{document}